\documentclass[pdflatex,sn-mathphys-num]{sn-jnl}

\usepackage{float}
\usepackage{graphicx}%
\usepackage{multirow}%
\usepackage{amsmath,amssymb,amsfonts}%
\usepackage{amsthm}%
\usepackage{mathrsfs}%
\usepackage[title]{appendix}%
\usepackage{xcolor}%
\usepackage{textcomp}%
\usepackage{manyfoot}%
\usepackage{booktabs}%
\usepackage{algorithm}%
\usepackage{algorithmicx}%
\usepackage{algpseudocode}%
\usepackage{listings}%
\usepackage{comment}
\usepackage{multirow}
\usepackage{natbib}
\usepackage{siunitx}    
\usepackage{rotating}   
\usepackage{tabularx}   

\theoremstyle{thmstyleone}%
\theoremstyle{thmstyletwo}%

\theoremstyle{thmstylethree}%

\begin{document}

\title[Article Title]{Application of the Brain Drain Optimization Algorithm to the N-Queens Problem}

\author*[1]{\fnm{Sahar} \sur{Ramezani Jolfaei}}\email{6709190@studenti.unige.it}

\author[2]{\fnm{Sepehr} \sur{Khodadadi HosseinAbadi}}\email{6660699@studenti.unige.it}

\affil*[1]{\orgdiv{Department of Mathematical, Physical and Natural Sciences}, \orgname{University of Genova}, \orgaddress{\city{Genova}, \country{Italy}}}

\affil[2]{\orgdiv{Department of Mathematical, Physical and Natural Sciences}, \orgname{University of Genova}, \orgaddress{\city{Genova}, \country{Italy}}}


\abstract{This paper introduces the application of the BRADO\footnote{Brain Drain Optimization} algorithm—a swarm-based metaheuristic inspired by the emigration of intellectual elites to the N-Queens problem. The N-Queens problem, a classic combinatorial optimization problem, serves as a challenge for applying the BRADO. A designed cost function guides the search, and the configurations are tuned using a TOPSIS-based\footnote{Technique for Order Preference by Similarity to Ideal Solution} multicriteria decision making process. BRADO consistently outperforms alternatives in terms of solution quality, achieving fewer threats and better objective function values. To assess BRADO's efficacy, it is bench-marked against several established metaheuristic algorithms, including Particle Swarm Optimization (PSO), Genetic Algorithm (GA), Imperialist Competitive Algorithm (ICA), Iterated Local Search (ILS), and basic Local Search (LS). The study highlights BRADO’s potential as a general-purpose solver for combinatorial problems, opening pathways for future applications in other domains of artificial intelligence.}




\maketitle

\section{Introduction}
\label{Introduction}
The N-Queens Problem is a puzzle in the realm of combinatorial optimization, intriguing researchers for generations. Imagine trying to place N queens on an N×N chessboard, ensuring no two queens can threaten each other by considering the chess rules\cite{N_Queens_Problem_Def}. Although this problem has a simple structure, it has been utilized to develop new problem-solving approaches. N-Queens Problem is often studied as a ‘mathematical recreation’, but it has several real-world applications such as multiprocessor scheduling under timing constraints\cite{N_Queens_Applications_4}, load balancing \cite{N_Queens_Applications_5}, parallel memory storage schemes, VLSI testing, traffic control and deadlock prevention \cite{N_Queens_Applications_6}, parallel memory, complete mapping problems, constraint satisfaction, and other physics, computer science and industrial applications\cite{N_Queens_Applications_1}; \cite{N_Queens_Applications_2}; \cite{N_Queens_Applications_3}.

The complexity of the N-Queens problem is one of its most interesting aspects. As N increases, the number of possible sets of queens on the board increases factorially. Specifically, the number of potential queen placements is $N^N$, making an exhaustive search infeasible for a large N\cite{N_power_of_N}. The N-Queens Problem is classified as NP-hard, which means that there is no known polynomial-time algorithm that can solve all instances of the problem efficiently\cite{mandziuk2002neural}. Various algorithms have been developed to solve the N-Queens problem. Some of them are mentioned as follows:
A systematic method for exploring possible configurations is backtracking algorithm. It incrementally builds solutions and abandons configurations that violate constraints\cite{thada2014performance}. Moreover, techniques like Simulated Annealing, Genetic Algorithm, and Tabu Search explore the solution space by moving from one potential solution to a nearby one, seeking improvements iteratively\cite{sadiq2010proposal}. Genetic algorithms are also powerful and simulate natural selection processes to evolve better solutions over generations. They use operations like selection, crossover, and mutation to explore the solution space\cite{saffarzadeh2010hybrid}. Also, there is constraint satisfaction programming, which involves formulating the problem with constraints and using constraint solvers to find solutions. It leverages the power of constraint propagation to efficiently prune the search space\cite{ayub2017exhaustive}.

The primary advantage of metaheuristics over the previous methods lies in their capability to handle large-scale instances within a reasonable time.\cite{yang2010nature}. As the n-queens problem is NP-hard, metaheuristic techniques are well-suited for solving it. Numerous studies have explored the use of metaheuristics for this problem, with methods such as Simulated Annealing (SA) being among them. \cite{tambouratzis1997simulated}; \cite{dirakkhunakon2009simulated}, Tabu Search (TS) \cite{martinjak2007comparison}, Genetic Algorithms (GA) \cite{homaifar1992nqueens}, Differential Evolution Algorithm (DEA) \cite{draa2010quantum}, and Ant Colony Optimization (ACO) \cite{khan2009solution}. Among these approaches, the Brain Drain Optimization (BRADO) algorithm has demonstrated remarkable effectiveness. Previous works have highlighted the versatility and effectiveness of the BRADO algorithm. For instance, they showcased BRADO's ability to outperform traditional algorithms like Particle Swarm Optimization (PSO) and Genetic Algorithms (GA) in avoiding local minima and handling high-dimensional problems.\cite{basiri2014introducing} 
Another study demonstrated BRADO's superiority in forming collaborative teams with minimal communication costs, outperforming other notable algorithms on datasets like DBLP and IMDb.\cite{BRADO2} Further exploring its capabilities, illustrated BRADO's efficiency in minimizing workflow makespan in cloud computing environments, a difficult NP-hard problem.\cite{BRADO4} Additionally, a new classifier inspired by BRADO was proposed, showcasing its flexibility in classification tasks across various domains.\cite{BRADO5}

Despite all the solutions proposed for the N-Queens problem, BRADO has not yet been tested on the N-Queens Problem. BRADO has been successful in some other NP-hard problems, and stands as a promising candidate to provide an efficient solution for the N-Queens Problem. This article explores the application of the BRADO algorithm to the N-Queens Problem, comparing its performance with well-known algorithms like Particle Swarm Optimization (PSO), Local Search (LS), Genetic Algorithms (GA), Imperialist Competitive Algorithm (ICA), Iterated Local Search (ILS), and Memetic Local Search (MLS). Our findings reveal that BRADO not only competes effectively but frequently surpasses other algorithms in discovering optimal solutions. The upcoming sections are organized as follows. Sect. 2 demonstrates how the N-Queens problem is solved using BRADO. In Sect. 3 the results are compared with well-known algorithms and finally, Sect. 4 contains the conclusion.

\section{Methodology}
The N-Queens problem was formulated as a combinatorial optimization problem, where the objective is to minimize the number of conflicts among queens placed on a $N \times N$ chessboard. The cost function evaluates candidate solutions by counting the number of attacking queen pairs, thereby guiding the search process toward non-conflicting arrangements. A pseudo-code for the cost function is shown as below:

\begin{algorithm}[H]
\caption{N-Queen Cost Function}
\begin{algorithmic}[1]
\Require An array \texttt{individual} representing queen positions
\Ensure Number of clashes (queens attacking each other)
\State Set \texttt{clashes} to 0
\State Compute \texttt{row\_col\_clashes} as the absolute difference between the length of \texttt{individual} and the number of unique elements in \texttt{individual}
\State Increment \texttt{clashes} by \texttt{row\_col\_clashes}
\For{each index $i$ from $1$ to length of \texttt{individual}}
    \For{each index $j$ from $1$ to length of \texttt{individual}}
        \If{$i \neq j$}
            \State Compute $dx = |i - j|$
            \State Compute $dy = |\texttt{individual}[i] - \texttt{individual}[j]|$
            \If{$dx = dy$}
                \State Increment \texttt{clashes} by $1$
            \EndIf
        \EndIf
    \EndFor
\EndFor
\State \Return \texttt{clashes}
\end{algorithmic}
\end{algorithm}

Since the BRADO algorithm was originally designed to optimize benchmark functions such as Sphere\cite{Sphere}, we adapted it to address the discrete nature of the N-Queens problem, in contrast to the continuous setting of those functions.

Upon completion of all runs, cost values and function evaluation counts were aggregated across configurations and problem instances. These metrics were then normalized and used as input to a TOPSIS-based multi-criteria decision-making process.\cite{TOPSIS} The purpose of this step was to systematically identify the most effective configuration for each instance by considering trade-offs between solution quality and computational effort.

Once optimal configurations were identified through the TOPSIS phase, they were re-evaluated independently. For each selected configuration, the corresponding problem instance was solved 5 additional times to validate the consistency and stability of the algorithm’s performance under the tuned parameters. This ensured that the reported results were not only derived from well-tuned algorithmic parameters but also validated under repeated and independent execution scenarios, thus reinforcing the reliability of the findings. BRADO algorithm was compared with the methods according to Table~\ref{tab:algorithm_types}:
\begin{table}[!htb] 
    \centering 
    \begin{tabular}{l c} 
        \hline
            \textbf{Algorithm Name} & \textbf{Algorithm Type} \\ 
        \hline
            BRADO  & Population-based \\
            GA     & Population-based \\
            ICA    & Population-based \\
            PSO    & Population-based \\
            ILS    & Single-solution-based \\
            LS     & Single-solution-based \\
            MLS    & Single-solution-based \\
        \hline
    \end{tabular}
    \caption{Algorithms that are used to solve N-Queens Problem}
    \label{tab:algorithm_types}
\end{table}

\subsection{TOPSIS-Based Taguchi Method}
In the TOPSIS-based tuning approach, multiple values are considered for each parameter. If all possible parameter combinations were tested, the number of test cases would be excessively large due to the many possible configurations. The Taguchi method helps reduce the number of required test cases by selecting a fractional factorial design approach. For example, in the MLS algorithm, instead of testing 64 cases, only 16 cases are examined. A detailed example of this reduction can be seen in Table~\ref{tab:taguchi-reduction}\cite{TOPSIS}.

\begin{table}[!htb]
    \centering
    \caption{Initial Cases and Reduced Cases Using the Taguchi Method}
    \label{tab:taguchi-reduction}
    \begin{tabular}{|l|c|c|}
        \hline
        \textbf{Algorithm Name} & \textbf{Initial Cases} & \textbf{Cases Reduced by Taguchi} \\
        \hline
        BRADO & 32,768 & 32 \\
        GA    & 256    & 16 \\
        ICA   & 16,384 & 32 \\
        ILS   & 128  & 32 \\
        LS    & 16     & 16 \\
        MLS   & 64     & 16 \\
        PSO   & 256    & 16 \\
        \hline
    \end{tabular}
\end{table}
A set of accepted hyper-parameters\cite{PSOParam, SurveyParam} was initially used for each algorithm which is shown in supplementary information (Table S1). Then, hyperparameter tuning was done specifically for the N-Queens problem. The final parameter settings are provided in the Supplementary Information (Table S2-S10).

\section{Results}
After determining the optimal values for each parameter, the algorithms were executed 10 times using the optimized parameters. The best, worst, and average values of the objective function cost and NFE (Number of Function Evaluations) were recorded, as shown below. This table provides a concise summary; the complete table with more detailed data is available in the supplementary information (Table S11-S19).

\begin{table}[!htb]
    \centering
    \renewcommand{\arraystretch}{1.2}
    \caption{Comparative Execution Results for N-Queens Problem}
    \label{tab:execution_results_combined}
    \begin{tabular}{l l c c c c c c}
        \toprule
        & & \multicolumn{5}{c}{\textbf{Problem Size (N)}} \\
        \cmidrule(lr){3-7}
        \textbf{Algorithm} & \textbf{Metric} & \textbf{8} & \textbf{25} & \textbf{100} & \textbf{200} & \textbf{500} & \textbf{1000} \\
        \midrule
        \multirow{2}{*}{BRADO} 
        & Min Cost & \textbf{0} & \textbf{0} & \textbf{0} & \textbf{1} & \textbf{1} & \textbf{1} \\
        & Min NFE & 16.5K & 42.1K & 72.2K & 53.5K & 57.3K & 43.3K \\
        \midrule
        \multirow{2}{*}{GA}
        & Min Cost & \textbf{0} & 2 & 77 & 184 & 544 & 1292 \\
        & Min NFE &  7.9K & 9.0K & 1.6K & 7.2K & 6.9K & 4.2K \\
        \midrule
        \multirow{2}{*}{ICA}
        & Min Cost & \textbf{0} & 7 & 61 & 161 & 476 & 974 \\
        & Min NFE & 8.3K & 9.8K & 24.7K & 35.7K & 26.5K & 18.2K \\
        \midrule
        \multirow{2}{*}{ILS}
        & Min Cost & 1 & 10 & 87 & 215 & 710 & 1477 \\
        & Min NFE & 1.7K & 3.3K & 6.6K & 6.7K & 435 & \textbf{216} \\
        \midrule
        \multirow{2}{*}{LS}
        & Min Cost & 1 & 15 & 116 & 285 & 759 & 1572 \\
        & Min NFE & \textbf{159} & \textbf{190} & \textbf{31} & \textbf{35} & 40 & 39 \\
        \midrule
        \multirow{2}{*}{MLS}
        & Min Cost & \textbf{0} & 11 & 95 & 213 & 691 & 1525 \\
        & Min NFE & 3.7K & 4.3K & 4.4K & 1.8K & \textbf{214} & 383 \\
        \midrule
        \multirow{2}{*}{PSO}
        & Min Cost & 1 & 7 & 75 & 188 & 481 & 978 \\
        & Min NFE & 3.2K & 9.3K & 18.1K & 12.4K & 7.8K & 2.7K \\
        \bottomrule
    \end{tabular}
\end{table}

Based on the obtained data, it can be observed that the BRADO algorithm achieves higher accuracy in most cases. As a result, the number of objective function calls is higher, the number of threats is lower, and the minimum, average, and best values of the threats are better than the alternatives. However, it should be noted that due to the higher accuracy of the algorithm, its execution time has been higher compared to other algorithms.

\section{Conclusion}

As observed, the BRADO algorithm has achieved the best results in the N-Queens Problem by obtaining the lowest number of threats under comparable conditions with other algorithms. In this algorithm, threshold $\alpha$ plays a key role in the search process. Defined threshold levels in the brain allow it to accept other points once reaching a specific threshold. Similarly, when a threshold is exceeded, the brain migrates to explore other local spaces of the problem. This balance between exploration and exploitation is a feature not present in other population-based metaheuristic algorithms.

Population-based meta-heuristic algorithms typically focus more on exploration, while single-solution-based algorithms emphasize exploitation. The BRADO algorithm benefits from both approaches, as it allows local optimization while simultaneously separating groups from the main population to reach newer optimal points.

\section*{Declarations}

\textbf{Conflict of Interest}

The authors declare that they have no competing financial interests or personal relationships that could have appeared to influence the work reported in this paper.

\vspace{0.5em}
\textbf{Funding}

This research did not receive any specific grant from funding agencies in the public, commercial, or not-for-profit sectors.

\vspace{0.5em}
\textbf{Data Availability}

The data that support the findings of this study are available from the corresponding author upon request.

\vspace{0.5em}
\textbf{Author Contributions}

Sahar Ramezani Jolfaei\footnote{Corresponding author. Email: 6709190@studenti.unige.it} studied and implemented the BRADO algorithm for the N-Queens Problem. Sepehr Khodadadi Hossein Abadi contributed to the comparative analysis. Both authors participated in writing and revising the manuscript and approved the final version.
\bibliography{sn-bibliography}

\end{document}